\documentclass[10pt,journal,compsoc]{IEEEtran}
\usepackage{graphicx}
\usepackage{comment}
\usepackage{amsmath,amssymb} 
\usepackage{color}
\usepackage{array}
\usepackage{url} 
\usepackage[hyperfootnotes=false]{hyperref}       
\usepackage{tabularx}
\usepackage{enumitem}
\usepackage{setspace}
\usepackage{stfloats}
\newcolumntype{P}[1]{>{\centering\arraybackslash}p{#1}}
\usepackage{booktabs, multirow, array, makecell, caption}
\newcommand{\first}[1]{\textcolor{red}{{\textbf{#1}}}}
\newcommand{\second}[1]{\textcolor{blue}{{\underline{#1}}}}

\ifCLASSOPTIONcompsoc
  \usepackage[nocompress]{cite}
\else
  \usepackage{cite}
\fi
\ifCLASSINFOpdf
\else
\fi
\begin{document}

\title{Neural Maximum A Posteriori Estimation on Unpaired Data for Motion Deblurring}

\author{Youjian~Zhang,
        Chaoyue~Wang,
        Dacheng Tao
}

\IEEEtitleabstractindextext{%
\begin{abstract}
   Real-world dynamic scene deblurring has long been a challenging task since paired blurry-sharp training data is unavailable. Conventional Maximum A Posteriori estimation and deep learning-based deblurring methods are restricted by handcrafted priors and synthetic blurry-sharp training pairs respectively, thereby failing to generalize to real dynamic blurriness. To this end, we propose a Neural Maximum A Posteriori (NeurMAP) estimation framework for training neural networks to recover blind motion information and sharp content from unpaired data. The proposed NeruMAP consists of a motion estimation network and a deblurring network which are trained jointly to model the (re)blurring process (\textit{i.e.} likelihood function). Meanwhile, the motion estimation network is trained to explore the motion information in images by applying implicit dynamic motion prior, and in return enforces the deblurring network training (\textit{i.e.} providing sharp image prior). The proposed NeurMAP is an orthogonal approach to existing deblurring neural networks, and is the first framework that enables training image deblurring networks on unpaired datasets. Experiments demonstrate our superiority on both quantitative metrics and visual quality over state-of-the-art methods. Codes are available on \url{https://github.com/yjzhang96/NeurMAP-deblur}.  

\end{abstract}

}

\maketitle

\IEEEdisplaynontitleabstractindextext

%
\IEEEpeerreviewmaketitle

\IEEEraisesectionheading{\section{Introduction}\label{sec:introduction}}

%
%
%
%

\IEEEPARstart{D}{ynamic} scene deblurring, which aims to 
Dynamic scene deblurring, which aims to restore the sharp content from a real-world blurry image, is a classic low-level vision task and would benefit serials of perceiving tasks, \textit{e.g.} image classification and object detection. However, due to the blind and non-uniform nature of dynamic blur, it has long been a challenging research topic.

\begin{figure*}[ht]
    \centering \vspace{4mm}
    \includegraphics[width=\linewidth]{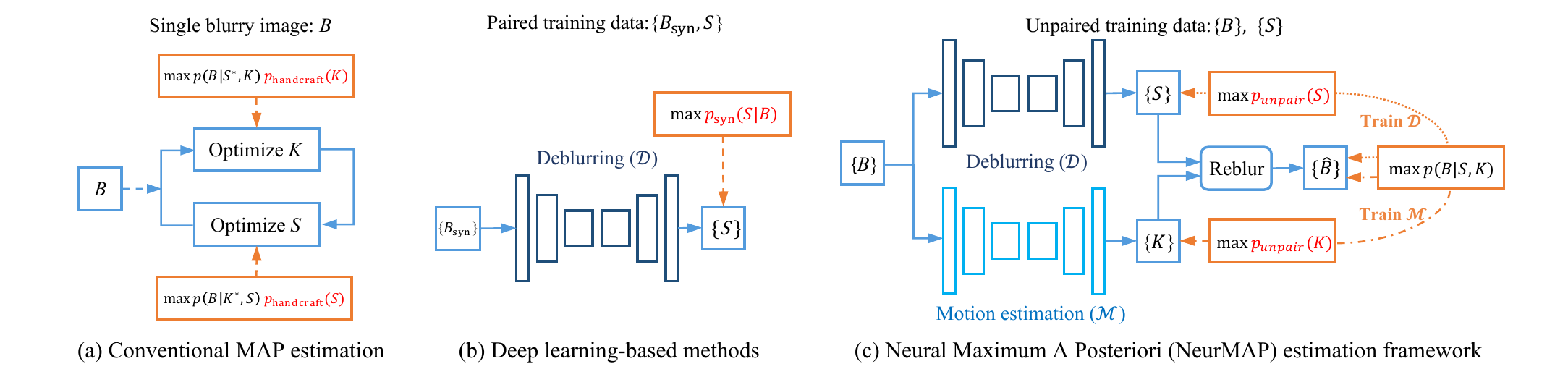}
    \caption{The schematic pipelines of (a) conventional MAP optimization; (b) previous deep deblurring methods; (c) the proposed NeurMAP framework. Unlike (a), neural networks presented in (b) and (c) are trained using a dataset.}
    \label{fig:pipeline}
\end{figure*}

Conventional deblurring methods usually treat deblurring as a maximum a posteriori (MAP) estimation problem~\cite{fergus2006removing,cho2009fast,jia2007single,levin2011efficient,hyun2013dynamic,pan2016blind}. Given a blurry image, the estimated blur kernels and the deblurred result are optimized alternately to approach the blurring process (Fig.~\ref{fig:pipeline} (a)). To reduce the intrinsic ill-posedness, different image priors and blur kernel priors~\cite{pan2016blind,whyte2012non,pan2016l_0,xu2010two,ren2020neural} are proposed. However, experiments show that the complex blurring process of real dynamic scenes cannot be accurately modeled by handcrafted priors. Moreover, performing MAP optimization on each blurry image is time-consuming and computationally inefficient, which further impedes these methods being widely used in real scenarios.

\begin{figure*}[ht]
    \centering
    \includegraphics[width=\linewidth]{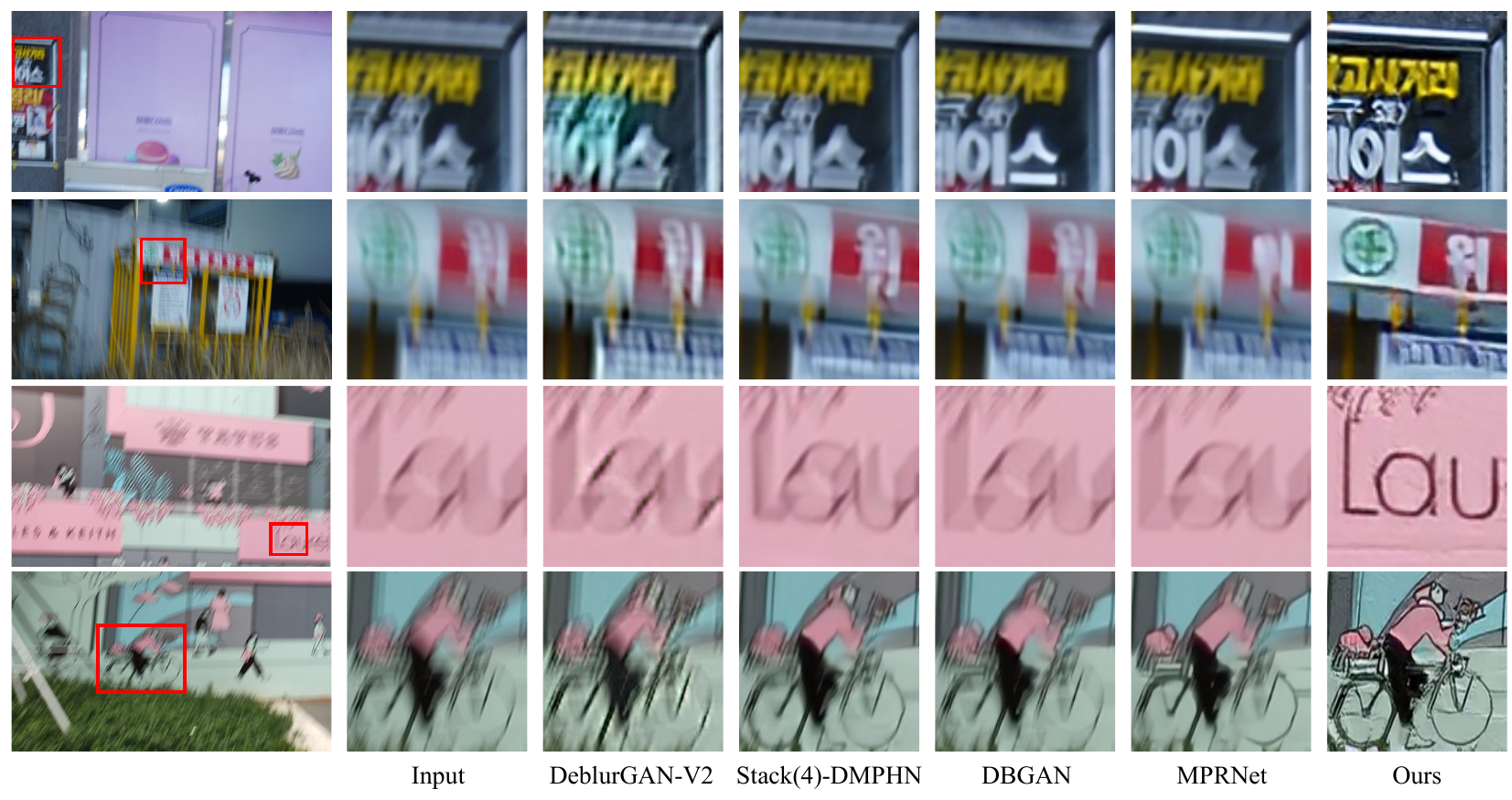}
    \caption{Comparisons on real-world image deblurring. The test image is sampled from both RealBlur dataset~\cite{rim2020real} and our collected Unpaired Real-world Blur (URB) dataset. Existing deep learning-based methods can only be trained on synthetic blurry-sharp pairs, thus failing to remove unseen real-world blur patterns. Our proposed NeurMAP successfully explored deblurring posterior probability from unpaired data and largely improved the deblurring results. }
    \label{fig:front-page}
\end{figure*}

Capturing real dynamic scenes with a high-frame-rate video camera, pseudo blurry images can be synthesized by averaging multiple adjacent frames~\cite{nah2017deep}. Meanwhile, the middle frame of the sequence is regraded as the sharp ground truth. As shown in Fig.~\ref{fig:pipeline} (b), with paired training data, deep neural networks~\cite{nah2017deep,tao2018scale,gao2019dynamic,zhang2019deep,suin2020spatially,purohit2020region,park2020multi} are trained to directly restore sharp content from blurry inputs. Taking advantages of the powerful learning capability, deep learning-based methods can achieve impressive dynamic scene deblurring results on test set and largely reduce the inference time. 
However, since domain gaps exist~\cite{zhang2020deblurring}, these well-trained networks may easily overfit synthetic blurry data and generalize poorly to unseen blurriness, especially real-world dynamic blurriness\footnote{Different from synthetic training pairs, current filming equipment cannot acquire paired real blurry images and sharp ground truth.}. As shown in Fig.~\ref{fig:front-page}, several most representative deep deblurring methods fail to recognize and handle real-world blurry images. 

In this paper, we aim to explore useful information contained in unpaired blurry and sharp images, then utilize the learned knowledge to improve deblurring performance. Inspired by conventional MAP estimation, a Neural Maximum A Posteriori (NeurMAP) estimation framework is proposed to explore the accurate deblurring posterior $p(S,K|B)$ from unpaired data, where $B$, $S$, $K$ represent the blurry image, the sharp image and blur kernels. As shown in Fig.~\ref{fig:pipeline} (c), we employ a deblurring network and a motion estimation network to recover the sharp content and blur kernels from blurry images. First, the deblurred result and the estimated blur kernels can be used to simulate the (re)blurring process, \textit{i.e.} modeling the likelihood probability $p(B|S,K)$.
Meanwhile, the motion estimation network is trained to predict dynamic motion information (\textit{i.e.} pixel-wise blur kernels) of an image. For example, the motion map estimated from a sharp image is penalized to be a zero filed, and we enforce the motion estimation network to continually explore unremoved blur/motions contained in deblurred images. In contrast, the deblurring network is trained to minimize the remaining blurriness explored by the motion estimation network. We show such an adversarial training process actually enforces the sharp image prior $p(S)$ and kernel prior $p(K)$ during training. 
These training objectives work together to maximum the posteriori probability over unpaired training data, \textit{i.e.} $p(S,K|B)\propto p(B|K,S)p(S)p(K)$. 

Overall, we make following contributions:
\begin{itemize}[]
\setlength{\itemsep}{1pt}
\setlength{\parsep}{0pt}
\setlength{\parskip}{0pt}
    \item A semi-supervised NeurMAP framework is proposed for training the deblurring model using unpaired data. Although synthetic blurry-sharp image pairs are provided for training stabilization, to our best knowledge, NeurMAP is the first deep framework that enables training image deblurring networks on unpaired data. 
    
    \item We build a directly connections between MAP estimation and training deblurring neural networks. Different from previous handcrafted priors and synthetic training pairs, the proposed NeurMAP explores deblurring posterior from unpaired data, which overcomes the overfitting problem on synthetic data. 
    
    
    
    \item Comprehensive experiments show that the proposed NeurMAP succeeds in modeling unseen blurring processes (both kernel-synthesized data and real-world data), and achieved significant improvement on real-world dynamic scene deblurring. In addition, even with only unpaired data, our NeurMAP is capable of recovering dynamic motions from a blurry image.   
    
    \item We discussed the effects of different generative adversarial networks (GANs) that employed in deblurring models, which further explain principles of adversarial training in the proposed NeurMAP framework. 
    
    
\end{itemize}

\section{Background}\label{sec:relatedwork}



\subsection{Conventional MAP-based Deblurring Methods}

Conventional MAP usually perform an optimization process on both sharp content and blur kernel iteratively. Since the optimization process is relatively fixed, the conventional MAP estimation methods focused on designing different priors to get a better deblurring performance. For example, Gaussian scale mixture priors \cite{fergus2006removing}, $l_1/l_2$-norms~\cite{krishnan2011blind}, $l_0$-norms~\cite{pan2016l_0}, total variation priors~\cite{xu2010two} and dark channel priors~\cite{pan2016blind} have been proposed during the past decade. Also, some specific camera motion constraints were introduced as regularization terms to shrink the potential solution space of blur kernels, such as ego motion~\cite{tai2010richardson}, 3D camera rotation~\cite{whyte2012non,whyte2014deblurring}, and camera forward~\cite{zheng2013forward}. Besides handcrafted priors, recently, \cite{li2018learning,ren2020neural,tran2021explore,dong2021learning} proposed to utilizing a neural network to explore 
kernel/image priors for MAP optimization, yet they still require an alternative optimization process. Also, none of them can explicitly represent the spatial-variant blur kernel in dynamic scenes.

Overall, MAP estimation build a theoretical foundation for image deblurring tasks, and achieved reasonable deblurring performance. However, most methods still rely on predefined priors or strong assumptions on the latent image or the causes of the motion blur. Thus, these methods usually fail in handling more complicated spatial-variant blur in real scenarios~\cite{nah2017deep}. Moreover, existing MAP-based methods have to perform the optimization on each blurry image, which could be time-consuming and inefficient.

\subsection{Learning-based Deblurring Methods}

More recently, 
benefited from the release of synthetic blurry/sharp datasets~\cite{nah2017deep,su2017deep}, end-to-end deblurring networks~\cite{nah2017deep,tao2018scale,kupyn2018deblurgan,wang2019edvr,kaufman2020deblurring, zamir2021multi,zhang2020video} are trained directly to restore the latent sharp content from a blurry input. 
Among these methods, \cite{nah2017deep,tao2018scale,gao2019dynamic} adopted the multi-scale restoration strategy, they explored the different parameter sharing schemes and skip-connections. Moreover, a multi-input multi-output UNet (MIMO-UNet) \cite{cho2021rethinking} is proposed to compress the multi-scale strategy in a single UNet structure. Despite multi-scale, multi-temporal \cite{park2020multi}/multi-stage \cite{zamir2021multi} strategy has also been studied to boost the deblurring performance. \cite{zhang2018dynamic} 
attempted to learn a variant RNN to model spatial-variant blurs. \cite{suin2020spatially,purohit2020region,yuan2020efficient,purohit2021spatially} also designed a spatial-variant convolutional module for the non-uniform feature of blurry region. With the transformer-based methods starting to show its power in high-level vision tasks, several transformer-based methods \cite{chen2021pre,wang2021uformer,zamir2021restormer} have also been proposed to explore the benefit of non-local feature aggregate in image restoration tasks.
In addition to the methods which aim to find a strong and effective network architecture, \cite{kupyn2018deblurgan,kupyn2019deblurgan,ramakrishnan2017deep} combined deblurring with generative adversarial networks (GANs). For example, Kupyn \textit{et al.} proposed DeblurGANs \cite{kupyn2018deblurgan,kupyn2019deblurgan} that valid the effectiveness of different backbones and GANs. 

However, all these learning-based methods rely heavily on the paired training dataset, and usually fail to recognize unseen blur patterns such as the real-world blur. There are methods \cite{zhang2020deblurring,rim2020real} want to fix the generalization problem by obataining more realistic pseudo training pairs. For example, Zhang \textit{et al.} \cite{zhang2020deblurring} proposed a learning-to-Blur GAN (BGAN) for generating blurry image with more realistic blur. Rim \textit{et al.} \cite{rim2020real} collected an un-aligned real-world dataset in low-light environment with a well-designed dual-camera system. However, the acquisition process of these training pairs is complicated and time-consuming. Compared with \cite{zhang2020deblurring,rim2020real}, the proposed NeurMAP can be directly trained on easily acquired unpaired data, and demonstrated better performance on unseen blur.


\section{Method}

\subsection{Preliminaries: Image Blurring \& Deblurring}

Mathematically, an image blurring process can be formulated as~\cite{fergus2006removing}:
\vspace{-2mm}
\begin{equation}
\vspace{-1mm}
    B = K \ast S + noise,
\end{equation}
where $B$ and $S$ represent blurry and sharp images, $K$ is the blur kernel that records relative motions during the exposure period. As an inverse process, image deblurring aims to recover sharp content $S$ given a blurry image $B$. 

\noindent \textbf{Maximum
a posteriori probability (MAP) deblurring.} In Bayesian statistics~\cite{lee1989bayesian}, the deblurring task can be treated as a MAP estimation problem~\cite{fergus2006removing}.
Given the blurry image $B$, MAP estimation aims to retrieve both the latent sharp image $S$ and blur kernels $K$ to maximize the posterior distribution:
\begin{equation}\label{eq:map}
\vspace{-1mm}
    p(K,S|B) \propto p(B|K,S)p(S)p(K),
\end{equation}
where likelihood $p(B|K,S)$ measures the (re)blurring fidelity; $p(S)$ and $p(K)$ define the priors of sharp images and blur kernels, which is designed to reduce the ill-posedness and enforce the smoothness of deblurred images and blur kernels. 
As shown in Fig.~\ref{fig:pipeline} (a), in conventional MAP-based deblurring methods, given a blurry image $B$, the optimization process is performed to iteratively optimize its sharp content $S$ and blur kernels $K$. Specifically, the objective of optimizing $K$ is to maximize $p(B|K,S)p(K)$, while the objective of optimizing $S$ is to maximize $p(B|K,S)p(S)$. 
The whole optimization process needs only a blurry image, yet not its corresponding sharp ground truth. 

\noindent \textbf{Deep neural networks trained using paired data.} 
With the paired blurry/sharp dataset, many deep neural networks are trained directly to maximize the probability $p(S|B)$ over a dataset, as shown in Fig.~\ref{fig:pipeline} (b). Specifically, the deblurred results $\{\hat{S}\}$ is enforced to be identical to ground truth sharp images$\{S_i\}$. The well-trained neural network largely improved the deblurring efficiency, yet usually fails generalizing to unseen blur patterns. Apparently, this pipeline cannot handle the unpaired data.

\noindent \textbf{Training neural networks to maximize posteriori probability (NeurMAP).} 
To tackle problem of the absence of sharp ground truth, our proposed NeurMAP aims to simultaneously obtain the sharp image $S$ and blur kernels $K$ in an end-to-end manner.
Inspired by conventional MAP, the posteriori possibility is achieved by several training objectives. Differently, NeurMAP substitute all the calculation and estimation with differentiable neural networks (Fig. \ref{fig:pipeline} (c)). Also, we design multiple loss functions as learnable image priors/kernel priors, instead of pre-defined priors in conventional methods. 
Besides paired data, the proposed NeurMAP is capable of learning deblurring posterior from unpaired data. More details are illustrated in the following sections~\ref{sec:framework} and \ref{sec:losses}.

\begin{figure*}
    \centering
    \includegraphics[width=\linewidth]{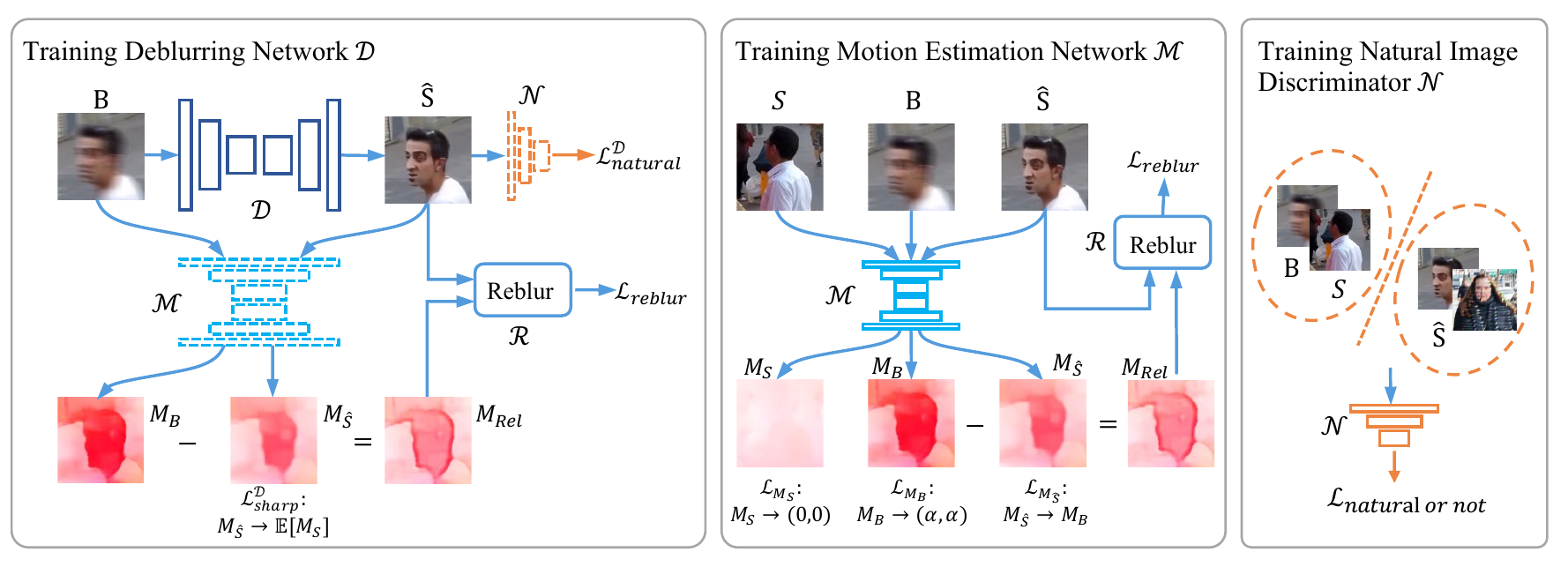}
    \caption{Our proposed framework consists of 3 trainable modules ($\mathcal{D}$, $\mathcal{M}$, $\mathcal{N}$) and one non-trainable reblurring module ($\mathcal{R}$). Specifically, $\mathcal{D}$ is trained with reblurring term and image prior term to generate a deblurred image. Meanwhile, $\mathcal{M}$ is trained to estimate the accurate motion maps of the images, and $\mathcal{N}$ is trained to distinguish whether a image is visually natural. In this figure, networks represented using dash line is fixed during training, networks with solid line is trained using corresponding losses, $\mathcal{L}_{(\cdot)}$. }
    \label{fig:network}
\end{figure*}

\subsection{The Proposed NeurMAP Framework}\label{sec:framework}

Overall, our framework consists three trainable neural networks: a deblurring network $\mathcal{D}$, a motion estimation network $\mathcal{M}$ and an natural image discriminator $\mathcal{N}$; and a non-trainable module: the reblurring module $\mathcal{R}$. In this section, we first introduce the functionality of these neural networks (module), our MAP-inspired training scheme is illustrated in Sec.~\ref{sec:losses}.

\noindent\textbf{Deblurring network $\mathcal{D}$}. Similar to existing deep deblurring methods, the proposed deblurring network 
$\mathcal{D}$ takes blurry images $B$ as inputs, and is trained to output the deblurred result $\hat{S} = \mathcal{D}(B)$. Our proposed NeurMAP is an orthogonal framework to existing deep neural networks. Therefore, the network $\mathcal{D}$ can adopt different architectures. In this paper, we valid the proposed NeurMAP on two representative deblurring backbone networks (section~\ref{sec:Experiments}).




\noindent\textbf{Motion estimation network $\mathcal{M}$}. Given an input image such as $B$, $\hat{S}$ or $S$, the motion estimation network $\mathcal{M}$ is trained to recover the motion information contained in the image, \textit{i.e.} how the image is blurred. 
Since the dynamic blur is usually spatial-variant, $\mathcal{M}$ predicts a motion map with the same resolution as its input. Similar to \cite{gong2017motion}, the blur kernel under the linear assumption can be described as a 2-dimensional motion vector. In our NeurMAP, the motion map can be written as $M(p) = (u_p,v_p)$, where $u_p$ and $v_p$ represent the horizontal and vertical components of the motion at the coordinate $p$. 
Through training $\mathcal{M}$, we aim to get an accurate motion representation from the blurry image. 
There are two benefits of learning an accurate motion representation in our method: i) such motion representation can be applied as a strong image prior for sharp image (sharp image tends to have small blur kernel) to enforce the deblurring process; ii) the relative difference between a blurry image $B$ and its deblurred result $\hat{S}$ can be utilized to reblur $\hat{S}$, \textit{i.e.} $M_{Rel}$ in Fig.~\ref{fig:network}, thereby optimizing the fidelity term $p(B|K,S)$.

\noindent\textbf{Non-trainable reblurring module $\mathcal{R}$}. Differentiable (re)blurring is stuided in serveral works~\cite{chen2018reblur2deblur,brooks2019learning,liu2020self,zhang2021exposure}.
Here, we follow the blur creation module proposed in \cite{zhang2021exposure}, since it is computationally efficient and easy to implement. Specifically, the relative motion vector $M(p)$ is discretized to $N$ steps and then utilized for warping the deblurred/sharp image. Taking the reblurring process as an example,
\begin{equation}\label{eq:reblur}
\small
    \hat{B} = \mathcal{R}(\hat{S},M_{rel}) =  \frac{1}{N} \sum_{n=0}^{N-1} \mathcal{W}(\hat{S},\frac{n}{N-1}(M_{rel}) - \frac{M_{rel}}{2}).
\end{equation}
$\mathcal{W}(\cdot,\cdot)$ is a warping function, which can help us obtain $N$ discretized frames during the blurring process. Given a relatively large $N$, the average of these discretized frames approaches the real blurring process, \textit{i.e.} $\hat{B} = K \ast \hat{S}$. 

\noindent\textbf{Natural image discriminator $\mathcal{N}$}. Considering the ill-posedness of the restored sharp image and blur kernel, the sharp image prior imposed by motion estimation network is not enough to regularize the restored image to be visually realistic. In fact, our experiments show this training using unpaired data may easily lead to unnatural artifacts. So we employ an natural image discriminator to regularize the deblurred results so that they are in the natural image domain. Training the natural image discriminator is equivalent to learn and optimize a image prior term.
Similar to DeblurGAN (v2)~\cite{kupyn2018deblurgan,kupyn2019deblurgan}, 
the proposed $\mathcal{N}$ adopts the PatchGAN structure~\cite{isola2017image} and aims to enforce deblurred images to have the same prior distribution with real-world images. Differently, the natural image discriminator only distinguishes the natural images (both blurry and sharp images) from generated ones (deblurred results are regraded as fake images). 

\subsection{NeurMAP Training Scheme}\label{sec:losses}

In this section, we introduce (i) the losses for training aforementioned networks, and (ii) how the training enforces maximizing a posteriori probability. Similar to conventional MAP-based deblurring methods, the proposed training scheme of NeurMAP aims to maximize $p(K,S|B)$ (Eq.~\ref{eq:map}). Differently, as shown in Fig.~\ref{fig:network}, our NeurMAP is trained on the whole training dataset (including both paired and unpaired data). 

\noindent\textbf{Data term: $p(B|\hat{S}, K)$}. Giveing the estimated blur kernels and deblurred results, the data term models the reblurring process, \textit{i.e.} maximizes likelihood probability $p(B|\hat{S}, K)$. In the proposed NeurMAP, we calculate the reconstruction losses between a reblurred image $\hat{B}$ with its blurry input $B$,
\begin{equation}
    \mathcal{L}_{reblur} = || B - \hat{B} ||^2 = || B - \mathcal{R}(\hat{S}, M_{rel})||^2,
\end{equation}
where $M_{rel}$ denotes the relative motion map. Since the reblurring module $\mathcal{R}$ is fully-differentable, the data term $\mathcal{L}_{reblur}$ is used to optimize both deblurring network $\mathcal{D}$ and motion estimation network $\mathcal{M}$, which is essential for maintaining content consistency.

\noindent\textbf{Kernel prior term: $p(K)$}.
Since we do not have the ground truth for neither deblurred outputs nor estimated motion maps, the model only trained with the data term will soon converge to a trivial solution, \textit{e.g.}  the deblurred $\hat{S}$ is identical with $B$ and the estimated motion is always zero. 
Enforcing kernel priors contributes to getting rid of such cases. Specifically, we concluded following implicit priors: (i) The motions $M_{B}$ estimated from a blurry image should be relatively large. We enforce $M_{B}(p) \to (\alpha,\alpha)$, where the $\alpha$ is regraded as the upper limit of the estimated motion. Considering the biggest blur kernel size in most previous methods~\cite{fergus2006removing,gong2017motion,sun2015learning} is around 30 to 40, in all of our experiments, we set $\alpha = 40$. 
(ii) The motion estimated from a sharp image should be a zero field, \textit{i.e.} we penalize the motion vector of each pixel estimated from a sharp image to be as close to zero as possible ($M_{S}(p) \to (0,0)$); 
(iii) During training, outputs of the deblurred network usually contains unremoved blurriness, and we hope the network $\mathcal{M}$ can explore those unremoved blurriness. It is reasonable to assume the unremoved motion is 
scaling down from the motion of blurry input, \textit{i.e.} $M_{\hat{S}}(p)\to M_{B}(p)$. 
Overall, these objectives can be written as:
\begin{equation}\label{eq:kernel prior}
    \small
  \begin{split}
  \mathcal{L}_{kernel}^{\mathcal{M}} &= \mathcal{L}_{M_{B}} + \mathcal{L}_{M_{S}} + \mathcal{L}_{M_{\hat{S}}} \\
  & =  \min_{\mathcal{M}} ||\mathcal{M}(B) - (\alpha,\alpha)|| +  ||\mathcal{M}(S)-(0,0)||   \\
  & \ \ \ \  + || \mathcal{M}(\mathcal{D}^*(B)) - detach(\mathcal{M}(B)) ||.
  \end{split}
\end{equation}
Here, $\mathcal{D}^* $ means the module $\mathcal{D}$ is fixed during back-propagation. The intention of these losses is to push the motion of the blurry image and the sharp image in two opposite directions, also to pull the motion of deblurred image to its blurry input. Together with other losses, multiple constraints work together to recover the accurate motion value.
In addition, since the blur kernels of dynamic blur are usually smooth along with the space, we apply the total variation (TV) regularization to encourage spatial smoothness of the estimated relative motion map:
\begin{equation}
\small
\begin{split}
\mathcal{L}_{tv}(M_{rel}) =  &\frac{1}{(w-1)h}\sum_{i=0}^{w-1}|M_{rel}(i,j)-M_{rel}(i+1,j)|	+ \\ &\frac{1}{w(h-1)}\sum_{j=0}^{h-1}|M_{rel}(i,j) - M_{rel}(i,j+1)|,
\end{split}
\end{equation}
where $(i,j)$ denotes the pixel location in a map with resolution of $w \times h$. $M_{\cdot}$ represents different motion maps estimated from images. 

\begin{figure}[t]
    \centering
    \includegraphics[width=\linewidth]{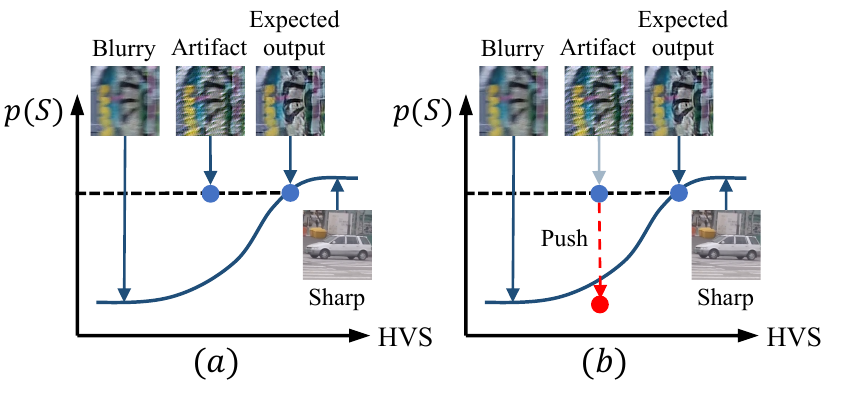}
    \caption{\textbf{Illustration of how our image priors work to eliminate artifacts.} The x-coordinate represents the evaluation metric by human visual system, and the y-coordinate represents the energy $E(I)$.  }
    
    \label{fig:discussion}
\end{figure}

\noindent\textbf{Image prior term: $p(\hat{S})$}. Since image deblurring is highly ill-posed, the regularization/priors on $\hat{S}$ are often the key to the success of deblurring. 
In our NeurMAP, two image priors, \textit{i.e.} sharp image prior and natural image prior, are learned and enforced by the motion estimation network and natural image discriminator, respectively.


Recall that the motion estimation network $\mathcal{M}$ is trained to estimate an accurate motion map from the input image. When training the deblurring network $\mathcal{D}$, we penalize the motion map estimated from the deblurred results to be as sharp as possible (sharp image prior), \textit{i.e.} $M_{\hat{S}}(p) \to \mathbb{E}[M_{S}]$:
\begin{equation}\label{eq:relative motion} 
\small
 \mathcal{L}_{sharp}^{\mathcal{D}} =  \min_{\mathcal{D}} ||\mathcal{M}^*(\mathcal{D}(B)) - \mathbb{E}[\mathcal{M}^*(S)] ||.
\end{equation}
We use the expectation $\mathbb{E}[\mathcal{M}^*(S)]$ since we do not have the sharp ground truth of the input blurry image, the expection value is regarded as the target motion value of all sharp images. Note that Eq.~(\ref{eq:relative motion}) works together with Eq.~(\ref{eq:kernel prior}) which training $\mathcal{M}$ to explore unremoved blur as much as possible, and these two losses form an adversarial training scheme.


The sharp image prior focuses on minimizing point-wise dense motions of the deblurred images. However, it may be cheated by artifacts in texture-level. Fig.~\ref{fig:discussion} gives a heuristic example. With only motion estimation network, results with artifacts (quite low visual quality) will also have a low training loss $\mathcal{L}_{sharp}^{\mathcal{D}}$. Therefore, 
we further introduce an natural image discriminator $\mathcal{N}$, which aims to distinguish whether a deblurred result is a natural image. Specifically, we adopt patchGAN~\cite{isola2017image} as our backbone and LSGAN's training losses~\cite{mao2017least}. We define natural images being with the numerical label `1', fake images being with the numerical label `0', then the adversarial losses is defined as:
\begin{equation}\label{eq:globalD}
\small
 \begin{split}
     \mathcal{L}^{\mathcal{N}}_{natural} = & \min_{\mathcal{N}} \mathbb{E}_{I\in\{B\}}[||\mathcal{N}(\mathcal{D}^*(I)) - 0||^2] \\
    & \ \ \  +\mathbb{E}_{I\in\{B,S\}}[||\mathcal{N}(I) - 1||^2] ,\\
     \mathcal{L}^{\mathcal{D}}_{natural} = & \min_{\mathcal{D}} \mathbb{E}_{I\in\{B\}}[||\mathcal{N}^{*}(\mathcal{D}(I)) - 1||^2].
 \end{split}
\end{equation}
It is worth noting that the natural image manifold includes sharp images and blurry images, $\{B,S\}$, since we observe that the deblurring process outputs blurry images in the beginning. As shown in our experiments, only using sharp images may lead to unstable training. 

Other than improving the naturalness, we apply a total variation (TV) loss to encourage the smoothness of the deblurred image. Since the sparse/uneven optimization on the motion map will cause the unsmoothness of deblurred image, we choose to smooth the motion map of the deblurred image $\hat{S}$:

\begin{equation}
\small
\begin{split}
\mathcal{L}_{tv}(M_{\hat{S}}) =  &\frac{1}{(w-1)h}\sum_{i=0}^{w-1}|M_{\hat{S}}(i,j)-M_{\hat{S}}(i+1,j)|	+ \\ &\frac{1}{w(h-1)}\sum_{j=0}^{h-1}|M_{\hat{S}}(i,j) - M_{\hat{S}}(i,j+1)|,
\end{split}
\end{equation}
where $(i,j)$ denotes the pixel location in a map with resolution of $w \times h$. $M_{\hat{S}}$ represents the motion map estimated from the deblurred image.

All these aforementioned image priors/regularizations is to encourage the deblurred results to be sharp and artifact-less.

\noindent\textbf{NeurMAP training scheme}. Till now, we introduced how to optimize each factor/term of the proposed NeurMAP framework. Before we conclude the overall training scheme, we hope to point that, in practical, our NeurMAP is trained in a semi-supervised manner. Similar to previous deep deblurring methods, the synthetic paired blurry/sharp images are employed to enforce the content loss, $\mathcal{L}_{content} = || S_{pair} - \mathcal{D}(B_{pair})||^2$, which enables our deblurring network $\mathcal{D}$ to have a baseline deblurring capability. Furthermore, we utilize the easily acquired unpaired data to train our NeurMAP framework, which can generalize deblurring to unseen blurring processes and achieve better deblurring results. Overall, 
in each training iteration, we apply content loss on paired data and aforementioned MAP losses on unpaired data. 

For the motion estimation network, its training loss is a weighted sum as:
\begin{equation}
\small
\begin{aligned}
    \mathcal{L}^{\mathcal{M}} &= \lambda\cdot\mathcal{L}_{reblur} +  \mathcal{L}_{kernel}^{\mathcal{M}} +  \mathcal{L}_{tv}(M_{rel}).
\end{aligned}
\end{equation}
In fact, through training $\mathcal{M}$, the data term and kernel prior term are optimized, \textit{i.e.} maximize $p(B|K,S)p(K)$, as shown in Fig.~\ref{fig:pipeline} (c).

For the deblurring network $\mathcal{D}$, the overall training loss is summarized as:
\begin{equation}\label{loss:train_D}
\small
    \mathcal{L}^{\mathcal{D}} = \lambda\cdot \mathcal{L}_{reblur} + \mathcal{L}_{sharp}^{\mathcal{D}} +  \beta \cdot \mathcal{L}_{natural}^{\mathcal{D}} + \mathcal{L}_{tv}(M_{\hat{S}}).
\end{equation}
Similarly, the training objective of $\mathcal{D}$ is equivalant to optimize data term and image prior term, \textit{i.e.} maximize $p(B|K,S)p(S)$.

Finally, the natural image discriminator $\mathcal{N}$ is trained,
\begin{equation}
\small
    \mathcal{L}^{\mathcal{N}} = \mathcal{L}^{\mathcal{N}}_{natural}.
\end{equation}
$\lambda$ and $\beta$ are only two hyperparameters that balance these training losses. For paired data, their content loss is weighted using $\lambda$. The detailed analysis of setting hyperparameters is provided in implementation details (section~\ref{subsec:ImplementationDetails}).


\begin{table*}
    \centering
    \footnotesize
    \renewcommand{\tabcolsep}{3.5pt} 
    \begin{tabular}{l c c c c c c c c c c}
    \toprule
    \multirow{2}{*}{Model} & \multicolumn{4}{c}{GoPro \cite{nah2017deep}}  & \multicolumn{4}{c}{Kernel-synthesized blur} & \multirow{2}{*}{Size (MB)} & \multirow{2}{*}{Runtime (s)}  \\
    \cmidrule(lr){2-5}
	\cmidrule(lr){6-9}	
     & PSNR$^\uparrow$ & SSIM$^\uparrow$ & LPIPS$_\downarrow$ & NIQE$_\downarrow$ & PSNR$^\uparrow$ & SSIM$^\uparrow$ & LPIPS$_\downarrow$ & NIQE$_\downarrow$ & & \\
    \midrule
    DeblurGAN-V2~\cite{kupyn2019deblurgan} & 29.55 & 0.9340 & 0.2535 & 5.290 & 21.61 & 0.7422 & 0.3339 & 3.802 & 60.9 & 0.124 \\
    Stack(4)-DMPHN~\cite{zhang2019deep} & 31.20 & 0.9453 & 0.1162 & 5.418 & 23.90 & 0.8332 & 0.2214 & 3.879 & 86.8 & 0.424  \\ 
    DBGAN~\cite{zhang2020deblurring} & 31.10 & 0.9424 & 0.1197 & 5.198 & 23.59 & 0.8278 & 0.2387 & 3.425 & \second{46.5} & 0.379\\
    MIMO-UNet+~\cite{cho2021rethinking} & \second{32.45} & 0.9572 & 0.0906 & 5.044 & 23.64 & 0.8484 & 0.2397 & 3.378 & 68.4 & \second{0.018} \\
    \midrule
    DMPHN~\cite{zhang2019deep} & 30.21 & 0.9345 & 0.1409 & 5.665 & 23.51 & 0.8187 & 0.2363 & 3.948 & 21.7 & 0.011 \\
    DMPHN + NeurMAP & 30.21 & 0.9389 & 0.1148 & \second{4.916} & \second{24.44} & \second{0.8500} & \second{0.1237} & \second{3.174} & \first{21.7} & \first{0.011} \\
    MPRNet~\cite{zamir2021multi} & \first{32.66} & \second{0.9591} & \second{0.0886} & 5.155 & 24.35 & 0.8499 & 0.2248 & 3.308 & 80.6 & 0.143\\    
    MPRNet + NeurMAP & 32.30 & \first{0.9601} & \first{0.0539} & \first{4.110} & \first{24.81} & \first{0.8629} & \first{0.1192} & \first{3.052} & 80.6 & 0.143\\
    
    \bottomrule
    \end{tabular}
    \caption{Quantitative comparison on both GoPro and kernel-synthesized blur datasets}
    \label{tab:quantitative_syn}
\end{table*}

\begin{figure*}[ht]
    \centering
    \includegraphics[width=\linewidth]{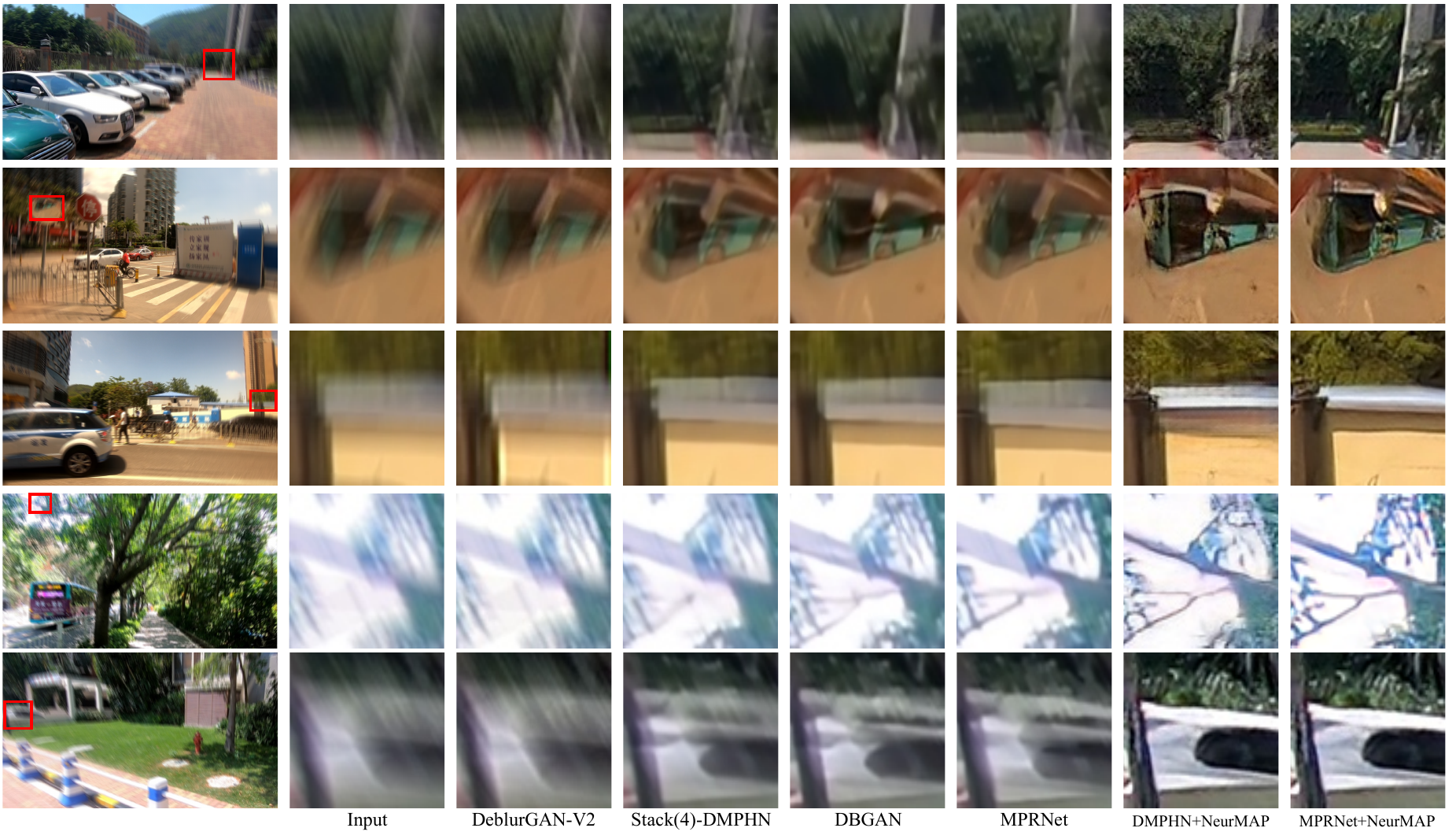}
    \caption{Visual comparison of the deblurring results on kernel-synthesized blur datasets. }
    \label{fig:compare_syn}
\end{figure*}

\section{Experiments}\label{sec:Experiments}

\subsection{Implementation Details}\label{subsec:ImplementationDetails}
Compared to previous deep deblurring methods, our proposed NeurMAP consists of several loss terms. To ensure the robustness of the proposed framework, we set hyperparameters in a heuristic way. Specifically, we first keep the coefficients for $\mathcal{L}_{kernel}$, $\mathcal{L}_{sharp}$ and $\mathcal{L}_{tv}$ the same, since all of them penalize on estimated motion maps. Our preliminary experiments illustrated that their weights have little influence on deblurring performance. Then, the weight of reblurring loss $\mathcal{L}_{reblur}$ and paired content loss $\mathcal{L}_{pair}$ are set as $\lambda$. We found that such reconstruction loss should be much larger than the other adversarial losses, where we set 100 times. Finally, the balance between the sharp image prior $\mathcal{L}_{sharp}^{\mathcal{D}}$ and the natural image prior $\mathcal{L}_{natural}^{\mathcal{D}}$ is controlled by $\beta$. Based on the experimental comparisons presented in Sec.~\ref{Sec:ablation}, $\beta$ is set to 0.1 to have a best trade-off between deblurring effect and eliminating artifact. 

In addition, we set $5\times10^{-5}, 10^{-4}$ and $10^{-4}$ as the initial learning rate for deblurring network, motion estimation network and natural image discriminator, respectively. All these learning rates linearly decayed to 0 at the end of the training. We use Adam \cite{kingma2014adam} solver for optimization, with $\beta_1=0.9$,
$\beta_{2}=0.999$ and $\epsilon=10^{-8}$. All discriminators are initialized using Xavier \cite{glorot2010understanding}, and bias is initialized to 0. 

\subsection{Datasets}
Similar to most learning-based methods, we employ GoPro dataset~\cite{nah2017deep} as our paired training data. In addition, we employ two kinds of unpaired data which have different types of blurring texture for testing the unpaired image deblurring ability. 

\noindent\textbf{Kernel-synthesized blur dataset}.
First, instead of the paired blur data created by averaging consecutive frames (GoPro dataset), our kernel-synthesized blur dataset is built by convolutional blur kernels. Specifically, we take the sharp images from dataset collected by \cite{gao2019dynamic}, then convolving them with the blur kernels synthesized by the method proposed in Gong \textit{et al.}~\cite{gong2017motion}. These blur kernels are represented as a spatial-variant 2D vector map. Finally, we applied 6 different blur kernel maps to each of 300 sharp images, and obtained 1800 blurry images. We splitted them to 5:1 for training/test set. In training phase, the paired information between blurry and sharp images are ignored. This dataset is mainly used for our quantitative evaluation since it has ground truth sharp images and corresponding blur kernels. 

\noindent\textbf{Real-world blur datasets.} Several real-world blur datasets are adopted to evaluate the effectiveness of unpair training on real-world blurry images. First, we have RealBlur dataset \cite{rim2020real} which shot blurry image and its unaligned sharp image through a dual-camera system. However, images in RealBlur dataset are shot under the low-light environment and by a professional equipment. To obtain a dataset that is closer to everyday life, we collect a real-world blur dataset with a handheld smartphone, and call it Unpaired Real-world Blur dataset (URB dataset). We film video clips of real dynamic scenes at 30 FPS, and extract every frame as blurry images. Meanwhile, we take multiple static photos for each scenes as sharp images. Unlike RealBlur dataset, all these blurry images taken in handheld smartphone have no groundtruth.
Our data collection process is conducted in a light-sufficient environment, and we try to use a relatively moderate movement speed. Finally, we build a dataset of total 3211 blurry images filmed from 24 different scenes.

\begin{figure*}[ht]
    \centering
    \includegraphics[width=\linewidth]{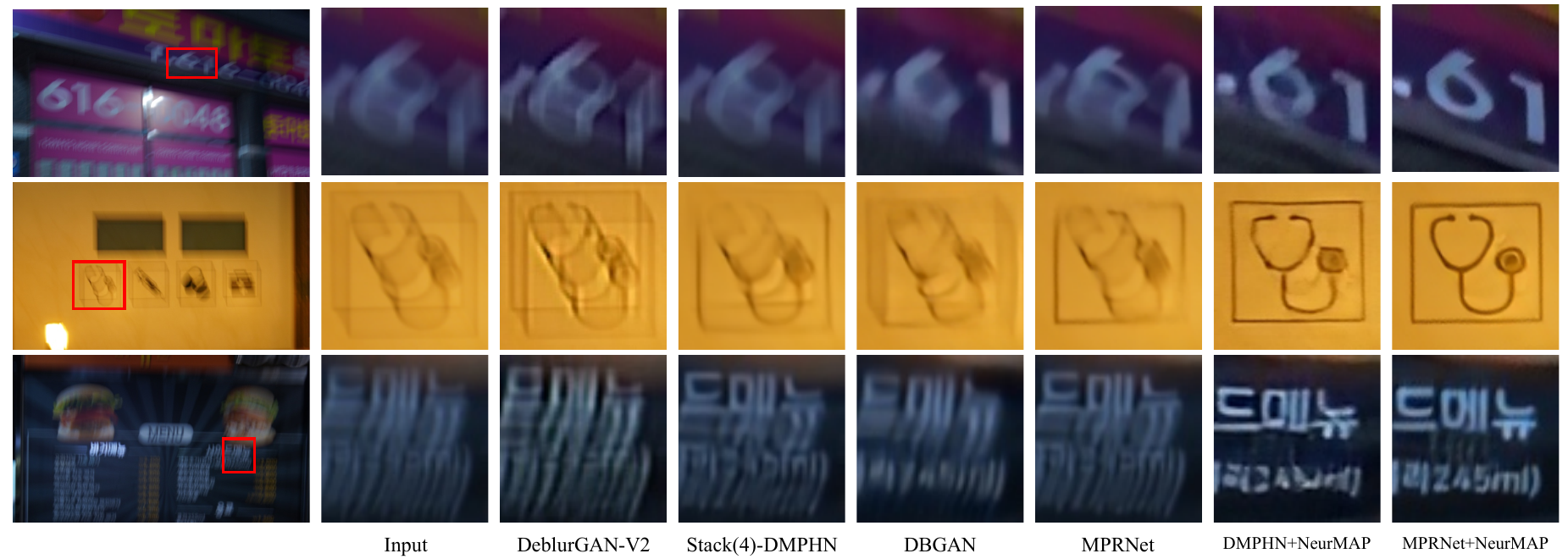}
    \caption{Visual comparison of the deblurring results on RealBlur datasets~\cite{rim2020real}.  }
    \label{fig:compare_RealBlur}
\end{figure*}

\subsection{Evaluation of Generalization Ability}
To evaluate the generalization ability, we employed two popular deblurring backbone networks, \textit{i.e.} DMPHN(-one stack) \cite{zhang2019deep} and MPRNet~\cite{zamir2021multi}, to test the proposed NeurMAP framework. Through comparing our models with several state-of-the-arts methods, we conduct both quantitative and qualitative comparisons on the GoPro dataset and the kernel-synthesized blur dataset since they provide ground truth sharp images. For RealBlur datasets, we adopt geometric alignment \cite{rim2020real} on deblurred images before conducting full-reference image quality assessment (PSNR, SSIM, LPIPS~\cite{zhang2018unreasonable}). Due to the lack of ground truth in URB dataset, we can only provide quantitative metric for no-reference image quality assessment (NIQE~\cite{mittal2012making}) and qualitative comparisons including user studies to valid our performance on real-world test set.

Among all comparison methods, the model proposed by DMPHN \cite{zhang2019deep}, MIMI-Unet~\cite{cho2021rethinking} and MPRNet~\cite{zamir2021multi} are trained with only content loss, while DeblurGAN-V2~\cite{kupyn2019deblurgan} and DBGAN \cite{zhang2020deblurring} are trained with a joint loss of content, perceptual and adversarial loss. 
In addition, DMPHN, DeblurGAN-V2, MIMI-Unet and MPRNet are trained only using GoPro dataset. Besides GoPro dataset, DBGAN utilized their proposed Real-World Blurry Image (RWBI) dataset to synthesize pseudo training pairs.
For a fair comparison, we use their official released well-trained models during inference.

\noindent\textbf{\small Comparisons on GoPro and kernel-synthesized blur datasets}.\\
As shown in Table~\ref{tab:quantitative_syn}, four metrics are utilized to conduct a quantitative comparison with existing methods for a comprehensive perspective. Among them, PSNR and SSIM mainly focus on the pixel values and structure similarity, while LPIPS~\cite{zhang2018unreasonable} and NIQE~\cite{mittal2012making} pay more attention to perceptual similarity and naturalness. As shown in Table~\ref{tab:quantitative_syn}, we can see that both the DMPHN~\cite{zhang2019deep} + NeurMAP and MPRNet~\cite{zamir2021multi} + NeurMAP
outperform all the state-of-the-art methods on the kernel-synthesized blur dataset in four metrics. Although the backbone model DMPHN has a inferior performance on GoPro dataset, its performance can exceed MPRNet by training with our NeurMAP framework. We can also observe that models like MIMO-Unet, though have a large improvement (2.24dB on PSNR) on GoPro dataset, affect little (0.13dB on PSNR) on unseen blur patterns, yet our method succeeds in generalizing the deblurring models to the domain of unseen blur pattern through training the unpaired data.

On the other hand, our method is shown to be very effective in improving LPIPS/NIQE metrics, as shown in kernel-synthesized blur dataset. The methods trained with NeurMAP can achieve much better score in LPIPS/NIQE compared to the method w/o NeurMAP, which suggests the results are visually clearer and closer to the sharp ground truth in terms of human visual perception. This inference can also be supported by the qualitative examples presented in the Fig.~\ref{fig:compare_syn}. Despite the fact that our method successfully recover the sharp content, we find that our results are significantly clearer on the edges,  and the details of our results are visually more abundant compared to others. As discussed in \cite{isola2017image}, our experiments also suggest that most existing deblurring models which trained with only the content loss would carry intrinsic blur brought by $L_1$/$L_2$ norm.


\begin{figure*}[ht]
    \centering
    \includegraphics[width=\linewidth]{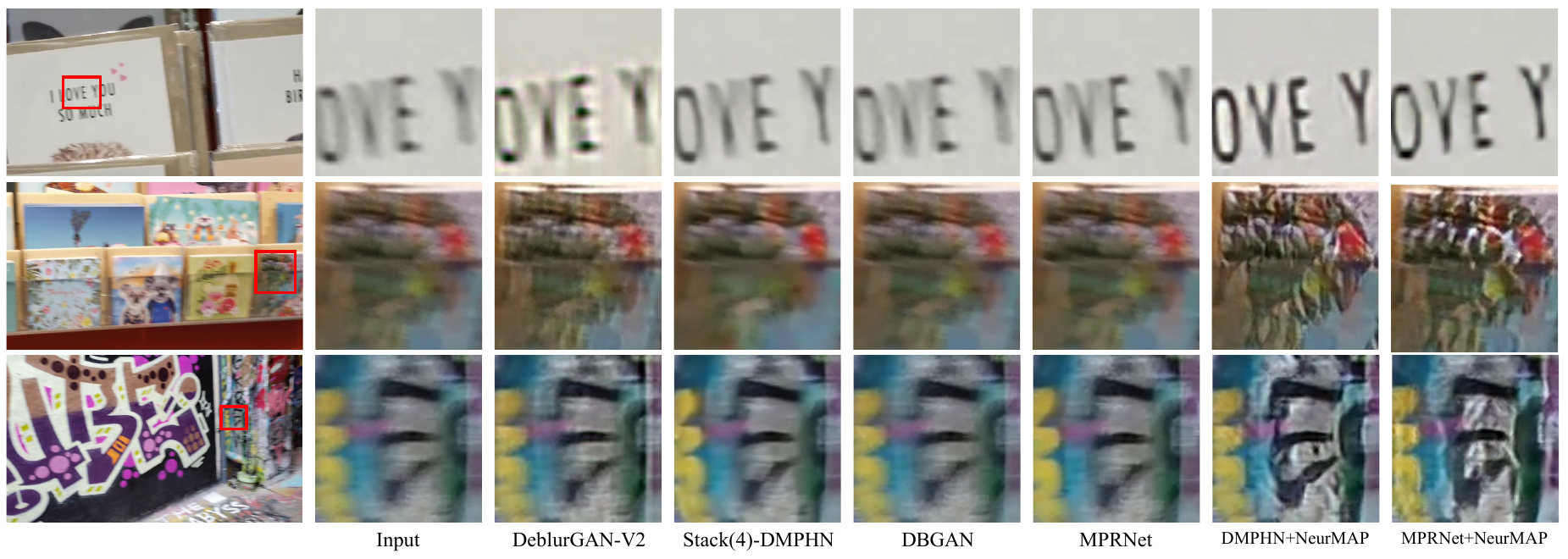}
    \caption{Visual comparison of the deblurring results on URB dataset.  }
    \label{fig:compare_URB}
\end{figure*}

\begin{figure*}
    \centering
    \includegraphics[width=\linewidth]{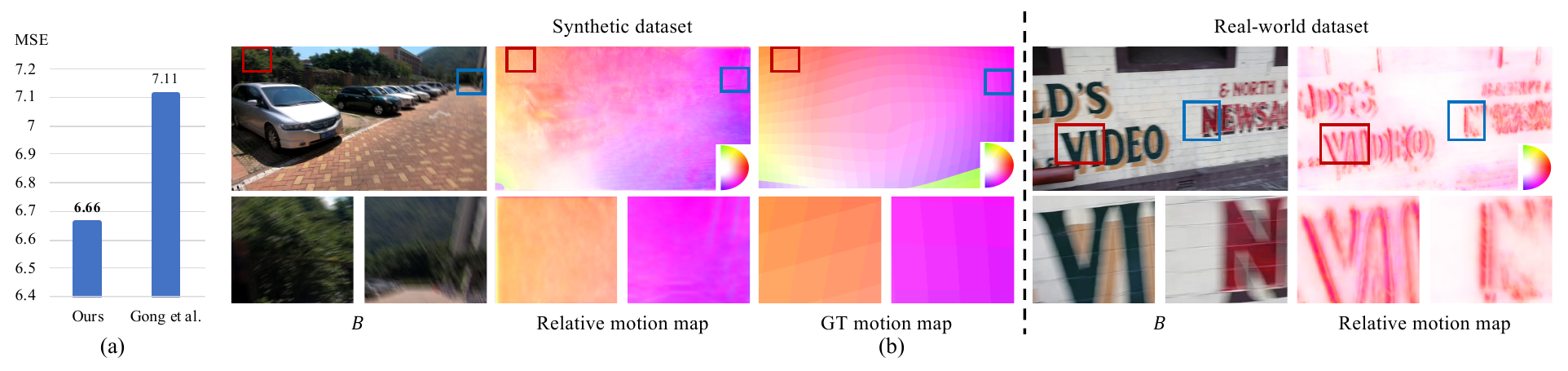} 
    \caption{Evaluation of the accuracy of estimated relative motion maps. (a) shows the result of MSE on kernel-synthesized blur dataset compared to Gong \textit{et al.}~\cite{gong2017motion}. (b) shows the visualization results on the kernel-synthesized blur dataset (left) and Real-world dataset (right). }
    \label{fig:vis_syn}
    
\end{figure*}

\begin{table}
    \centering
    \scriptsize
    \renewcommand{\tabcolsep}{3.5pt} 
    \begin{tabular}{l c c c c c }
    \toprule
    \multirow{2}{*}{Model} & \multicolumn{4}{c}{RealBlur \cite{rim2020real}}
    & URB \\
    \cmidrule{2-6}
    & PSNR & SSIM & LPIPS & NIQE & NIQE \\
    \midrule
    DeblurGAN-V2~\cite{kupyn2019deblurgan} & 26.52 & 0.8052 & 0.1935 & 5.464 & 4.590 \\
    Stack(4)-DMPHN~\cite{zhang2019deep} & 27.78 & {0.8469} & 0.1920 & 5.728 & 5.354 \\
    DBGAN~\cite{zhang2020deblurring} & 24.93 & 0.7451 & 0.2945 & 5.323 & 5.050 \\
    MIMO-UNet+~\cite{cho2021rethinking} & 27.63 & 0.8368 & 0.1991 & 5.442 & 4.818 \\
    \midrule
    DMPHN~\cite{zhang2019deep} & 27.52 & 0.8369 & 0.1973 & 5.682 & 5.201 \\
    DMPHN + NeurMAP & 28.05 & 0.8444 & \second{0.1457} & \first{4.140} & \first{4.302} \\
    MPRNet~\cite{zamir2021multi} & \first{28.70} & \first{0.8730} & 0.1527 & 5.308 & 5.002 \\
    MPRNet + NeurMAP & \second{28.49} & \second{0.8603} & \first{0.1425} & \second{4.834} & \second{4.570} \\
    \bottomrule
    \end{tabular}
    \caption{Quantitative comparison on real-world datasets}
    \label{tab:quantitative_real}
\end{table}

\noindent\textbf{Generalization ability on Real-world blur datasets}.\\
We perform NeurMAP training on unpaired RealBlur and URB dataset, and evaluate the generalization ability on the the test set. Quantitative comparisons can be found in Table \ref{tab:quantitative_real}. For RealBlur dataset, PSNR, SSIM and LPIPS metrics are calculated for geometrically aligned deblurred images. Similar to the analysis carried out from kernel-synthesized blur dataset, the proposed NeurMAP performs better in LPIPS/NIQE metrics compared to state-of-the-art methods. We should point out that, since our method do not rely on any regression-based loss (L1/L2 losses) which is favored by pixel value metrics, \textit{i.e.} PSNR/SSIM, it is very hard for our method to obtain a high PSNR/SSIM score. On the contrary, our method tends to synthesize misaligned high-frequency details which may be penalized by PSNR/SSIM \cite{saharia2021image,chen2018fsrnet,menon2020pulse}. Moreover, comparing the model trained with NeurMAP with model w/o NeurMAP, we can see the improvement on DMPHN is more obvious than that on MPRNet, we infer the MPRnet is harder to train for its heavy-weighted structure.

We can also observe a better visual quality results on real-world test data. As shown in Fig.~\ref{fig:compare_RealBlur}, Fig.~\ref{fig:compare_URB}, we provide examples from RealBlur and URB test set respectively. We can see nearly all the models trained on GoPro dataset fail to generalize to real-world blur. In fact, they tend to output an identical image when handling unseen blurriness. In addition, although DBGAN \cite{zhang2020deblurring} utilize real-world blurry images in RWBI dataset as training data, it also fails to recover sharp content from real-world blurry images.  

\begin{figure}[ht]
    \centering
    \includegraphics[width = 0.8 \linewidth]{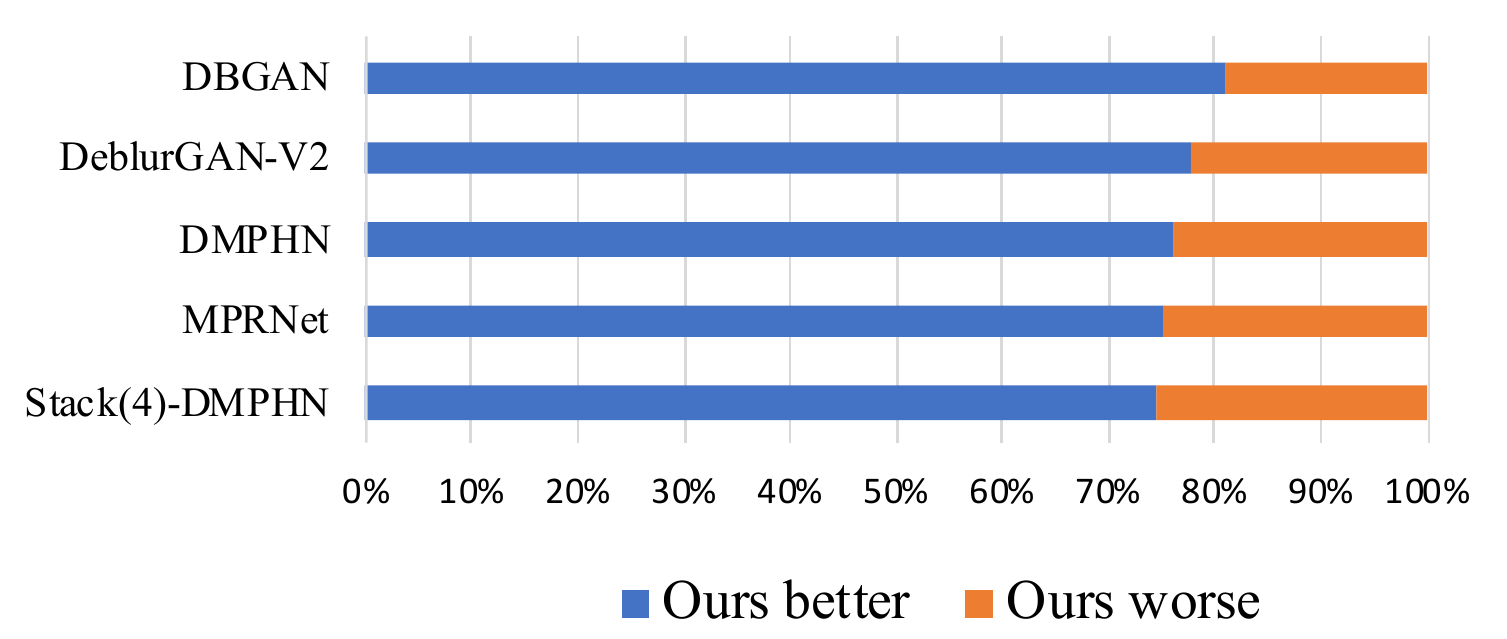}
    \caption{User preference compared to other methods.}
    \label{fig:User study}
\end{figure}

Due to the lack of quantitative metric to evaluate the performance of our model on the Real-world data without ground truth. A user study is conducted to validate the visual quality of our well-trained DMPHN + NeurMAP model(Fig.~\ref{fig:User study}). We selected 50 real-world blurry images, for each image, we asked 10 users on Amazon Mechanical Turk to compare the deblurring results of different methods.
The result shows that about 70\% to 80\% of people think our results are visually better than the others.

\begin{figure*}[ht]
    \centering
    \includegraphics[width=\linewidth]{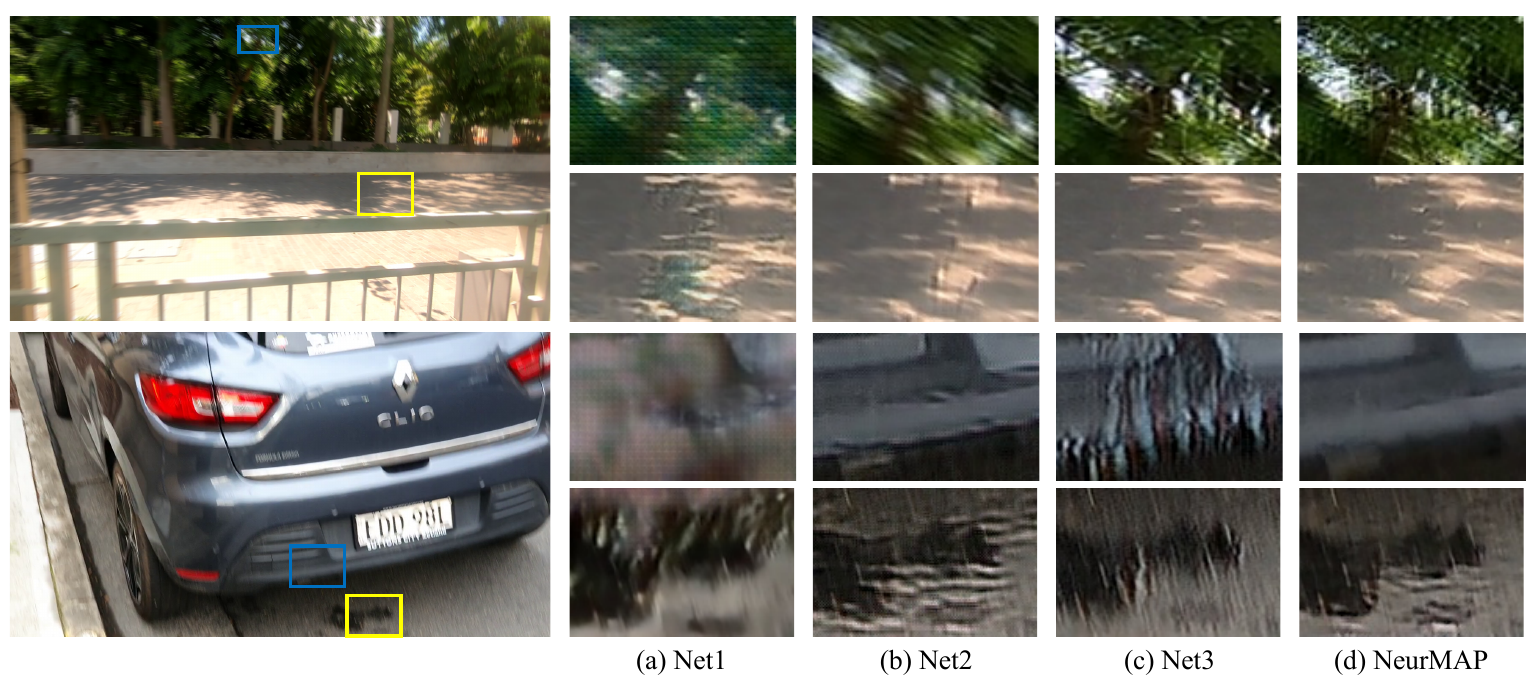}
    \caption{Qualitative results of different adversarial training schemes. (a) shows the content inconstancy brought by the absent of $\mathcal{M}$ and reblurring module. (b)-(d) demonstrate the effectiveness of $\mathcal{N}$ in eliminating artifacts.}
    \label{fig:compare_GAN}
\end{figure*}

\begin{table}[]
    \centering
    \footnotesize
    \begin{tabular}{l c c c c c}
    \toprule
        \multirow{2}{*}{Method} & \multirow{2}{*}{$\mathcal{M}+\mathcal{R}$} & \multicolumn{2}{c}{Discriminator} & \multirow{2}{*}{PSNR} & \multirow{2}{*}{SSIM}\\
         & & Sharp & Natural &  & \\
         \midrule
         Baseline &      -      & - &       -    & 23.51 & 0.8187 \\
         Net1 &            & \checkmark &           & 18.99 & 0.7035 \\
         Net2 & \checkmark &            &           & 21.94 & 0.7409 \\
         Net3 & \checkmark & \checkmark &           & 23.65 & 0.8280 \\
         NeurMAP & \checkmark &            & \checkmark & \textbf{24.44} & \textbf{0.8500} \\
         \bottomrule
    \end{tabular}
    \caption{Quantitative comparison of different training schemes.}
    \label{tab:training_scheme}
\end{table}

\begin{figure}
    \centering
    \includegraphics[width = \linewidth]{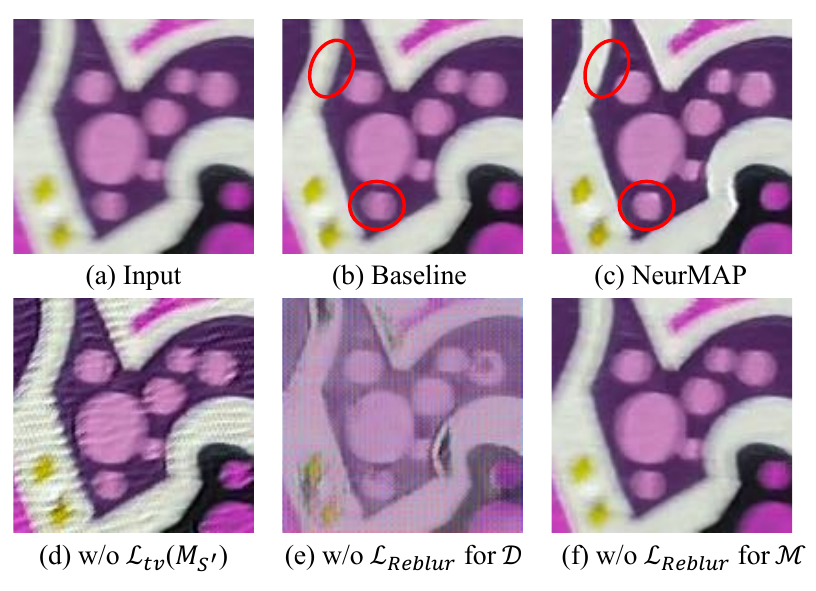}
    
    \caption{Ablation studies on TV and reblurring terms.}
    \label{fig:Ablation}
\end{figure}

\noindent\textbf{Analysis of the estimated relative motion map.} \\
Our framework performs the deblurring and motion estimation in a collaborative manner. The more accurate motion information is estimated, the better the deblurring performance will be achieved. Here, we validate that our well-trained motion estimation network can estimate accurate and dynamic motions/blur kernels in an explicit representation. 
First, we provide a quantitative evaluation on kernel-synthesized blur dataset since they have ground truth blur kernels. Specifically, we calculate the mean squared error (MSE) between estimated motion vectors and ground truth. Figure \ref{fig:vis_syn} (a) shows our estimated motion vectors are more accurate than a former supervised motion estimation method~\cite{gong2017motion}. Note that we follow \cite{gong2017motion} to build the kernel-synthesized blur dataset and our NeurMAP did not employ any paired information or ground truth motions during the training. Also, we visualize the relative motion map estimated from both kernel-synthesized blur dataset and real-world dataset as shown in fig.~\ref{fig:vis_syn} (b). Our estimated motion maps are highly consistent with the ground truth motion maps in both direction and magnitude. Although there is no ground truth for real-world test image, we observe the estimated motion maps mainly focus on the blurry edges, which partly explains how the motion estimation contributes to image deblurring.

\begin{figure*}
    \centering
    \includegraphics[width=\linewidth]{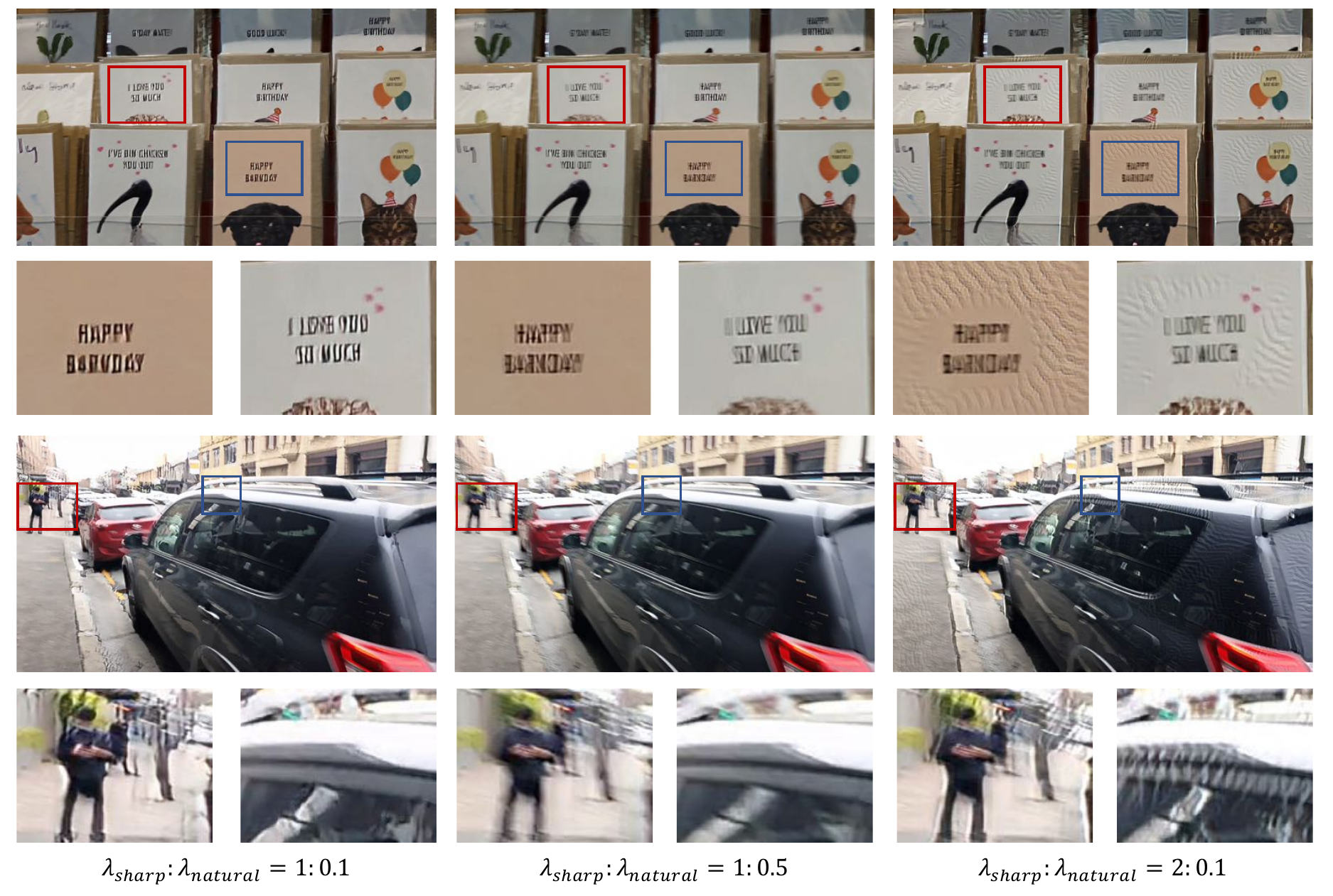}
    \caption{Results of different hyperparameters of $\mathcal{L}_{sharp}^{\mathcal{D}}$ and $\mathcal{L}_{natural}^{\mathcal{D}}$. From left to right shows 3 different settings of hyperparameters. The first one generates reasonable results; the second one tends to generate blurry output; the third one will cause more artifacts.  }
    \label{fig:different ratio}

\end{figure*}

\begin{table}
    \centering
    \footnotesize
    \begin{tabular}{l P{0.7cm} P{0.7cm} P{0.7cm} P{0.7cm} }
    \toprule
         Model (backbone -- DMPHN)& PSNR$^\uparrow$ & SSIM$^\uparrow$ & LPIPS$_\downarrow$ & NIQE$_\downarrow$ \\
         \midrule
        $\lambda_{sharp}:\lambda_{natural}=1:0.5$ & 22.16 & 0.7558 & 0.2205 & 3.493\\
         
         $\lambda_{sharp}:\lambda_{natural}=2:0.1$ & 21.42 & 0.7312 & 0.2336 & \textbf{3.095}\\
        
         $\lambda_{sharp}:\lambda_{natural}=1:0.1$ &  \textbf{24.44} & \textbf{0.8500} & \textbf{0.1237} & 3.174 \\
    \bottomrule
    \end{tabular}
    \caption{Quantitative comparison for different ratio of hyperparameters $\mathcal{L}_{sharp}^{\mathcal{D}}$ and $\mathcal{L}_{natural}^{\mathcal{D}}$.}
    \label{tab:hyperparameter}
\end{table}

\subsection{Evaluation of Training Strategies and Losses}\label{Sec:ablation}
To analyze the effectiveness of our proposed framework, we compare the deblurring networks trained using different strategies/losses. For fair comparison, all the comparison methods conduct a semi-supervised learning scheme. 

\noindent \textit{\textbf{A: Ablation studies on training strategies.}} \\
We design several training strategies to validate the effectiveness of our proposed modules. Quantitative results are evaluated with kernel-synthesized test data in DMPHN backbone ( Table~\ref{tab:training_scheme}). First of all, since a straightforward way of training on unpaired data is GAN-based unsupervised learning, our ablation model-\textbf{Net1} employs a vanilla sharp image discriminator which distinguishes between sharp and blurry images. The motion estimation network and corresponding reblurring module are removed. As shown in Table~\ref{tab:training_scheme}, this strategy causes a significantly drop in PSNR and SSIM scores since it cannot preserve the content. \textbf{Net2} introduces the motion estimation network but abandons image discriminators. Compared to Net1, the content is better preserved (better PSNR and SSIM). However, by observing deblurred samples 
in Fig.~\ref{fig:compare_GAN}, deblurred images suffer from the artifact caused by the sharp image prior. Compared to our NeurMAP, \textbf{Net3} employs a sharp image discriminator instead of our natural image discriminator. This combination improves the baseline models' PSNR and SSIM scores, yet two different sharp image priors bring instability to the training, and cannot handle artifacts well. Both quantitative and qualitative results are inferior to our NeurMAP. 

\noindent\textit{\textbf{B. Ablation studies on TV loss and reblurring loss.}} \\
In addition to validating the losses designed for each training module as a whole, we also provide visual comparisons of some specific losses. Fig.~\ref{fig:Ablation} (d) shows that without TV regularization, the texture artifacts will arise. Moreover, training deblurring network without reblurring loss will soon lose control of the content (Fig.~\ref{fig:Ablation} (e)). Finally, training motion estimation network without reblurring loss will cause all the motion maps to converge to 0, which fails to enforce the deblurring (Fig.~\ref{fig:Ablation} (f)).

\noindent\textit{\textbf{C. Different ratio of hyperparameters $\mathcal{L}_{sharp}^{\mathcal{D}}$ and $\mathcal{L}_{natural}^{\mathcal{D}}$} }
Losses $\mathcal{L}_{sharp}^{\mathcal{D}}$ and $\mathcal{L}_{natural}^{\mathcal{D}}$ are acted as sharp and natural image priors of training the deblurring network $\mathcal{D}$ respectively. Specifically, the training loss $\mathcal{L}_{sharp}^{\mathcal{D}}$ tends to make network $\mathcal{D}$ generate sharpened output, yet it may raise artifacts. On the other hand, the training loss $\mathcal{L}_{natural}^{\mathcal{D}}$ derived from natural image discriminator $\mathcal{N}$ forces $\mathcal{D}$ to eliminate artifacts while accepting blurry output. So there is a trade-off between the effect of $\mathcal{M}$ and $\mathcal{N}$, in order to generate sharpened and artifact-free output. As shown in Fig.~\ref{fig:different ratio}, we show the results come from different settings of hyperparameters of $\mathcal{L}_{sharp}^{\mathcal{D}}$ and $\mathcal{L}_{natural}^{\mathcal{D}}$.

We define the weight of $\mathcal{L}_{sharp}^{\mathcal{D}}$ and $\mathcal{L}_{natural}^{\mathcal{D}}$ in loss function Eq. 11 (main submission) as $\lambda_{sharp}$ and $\lambda_{natural}$, respectively. As we can see, $\lambda_{sharp} : \lambda_{natural} = 1:0.1$ generates reasonable deblurring results. When we raise $\lambda_{natural}$ to 0.5, there is little deblurring effect in the results. On the contrary, when we raise $\lambda_{sharp}$ to 2, the unsmooth artifact will arise in the deblurring results. The quantitative results on kernel-synthesized blur dataset (Table~\ref{tab:hyperparameter}) also show that all the other models have a poor performance on PSNR/SSIM, which means they fail in restoration fidelity. It is also interesting to observe that the kernel prior loss $\mathcal{L}_{sharp}^{\mathcal{D}}$ is beneficial to improve the NIQE metric, which shows that $\mathcal{L}_{sharp}^{\mathcal{D}}$ is effective in generating sharpened texture.

\subsection{Discussion and Limitation}
Learning to deblur from unpaired data is a challenging task. Here, we attempt to further clarify the principle and limitations of our NeurMAP through two questions. 

\noindent\textit{\small Q. Differences between GAN-based methods with NeurMAP?}

\noindent A: GANs are widely used in image processing tasks to improve the realistic of generated images. However, since GANs only penalize the similarity between two distributions (\textit{e.g.} $p(S)$ and $p(\hat{S})$), they cannot handle the unpaired training data without using other constraints. In the proposed NeurMAP, a reblurring term is presented to optimize the likelihood $p(B|K,S)$, which is critical to keep the content consistency and the physical meaning of kernels. 

\begin{figure}[t]
    \centering
    \includegraphics[width=\linewidth]{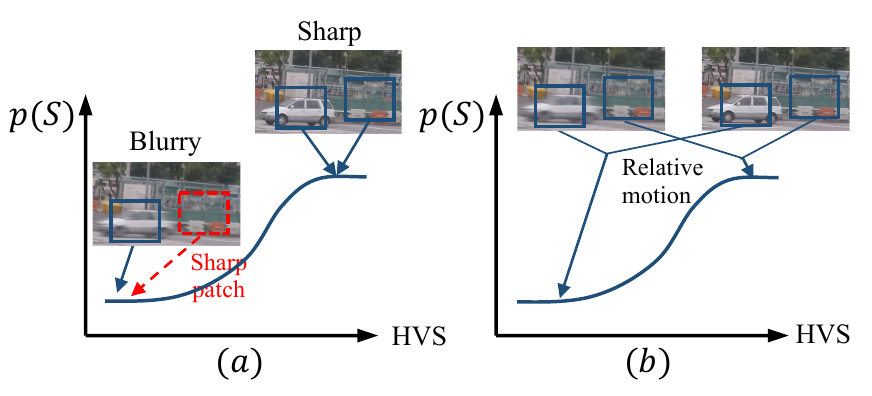}
    \caption{Illustration of relative motion map. The x-coordinate represents the evaluation metric by human visual system, and the y-coordinate represents the probability $p(S)$. (a)-(b) show the reason why we choose relative motion rather than absolute one as our reblurring motion.   }
    
    \label{fig:relative_motion}
\end{figure}

\noindent\textit{\small Q. Why we apply relative motion maps for reblurring?}\\
\noindent A: An alternative of using relative motion maps for reblurring is applying motion maps of blurry input (referred to as absolute motion maps) for reblurring. However, we find the relative motion maps are effective in solving the ambiguity problem of blurry/sharp regions.
As shown in Fig.~\ref{fig:relative_motion} (a), there always exists relative sharp patches within a blurry dataset. For a motion estimation network, it is ambiguous/impossible to distinguish sharp patches within a blurry image with patches in a sharp image. This may cause the motion estimated from some sharp region to be non-zero. However, with relative motion being calculated, the non-zero regions can be eliminated by the motion abstraction  (Fig.~\ref{fig:relative_motion} (b)). 
Similarly, for the natural image discriminator, directly push blurry/deblurred images $B/\hat{S}$ away from sharp images $S$ may lead to training instability. Therefore, introducing blurry images as natural images contribute to stable $\mathcal{N}$'s training.

\noindent\textit{\small Q. Why semi-supervised learning is performed? (Our limitation.)}

\begin{figure}
    \centering
    \includegraphics[width=\linewidth]{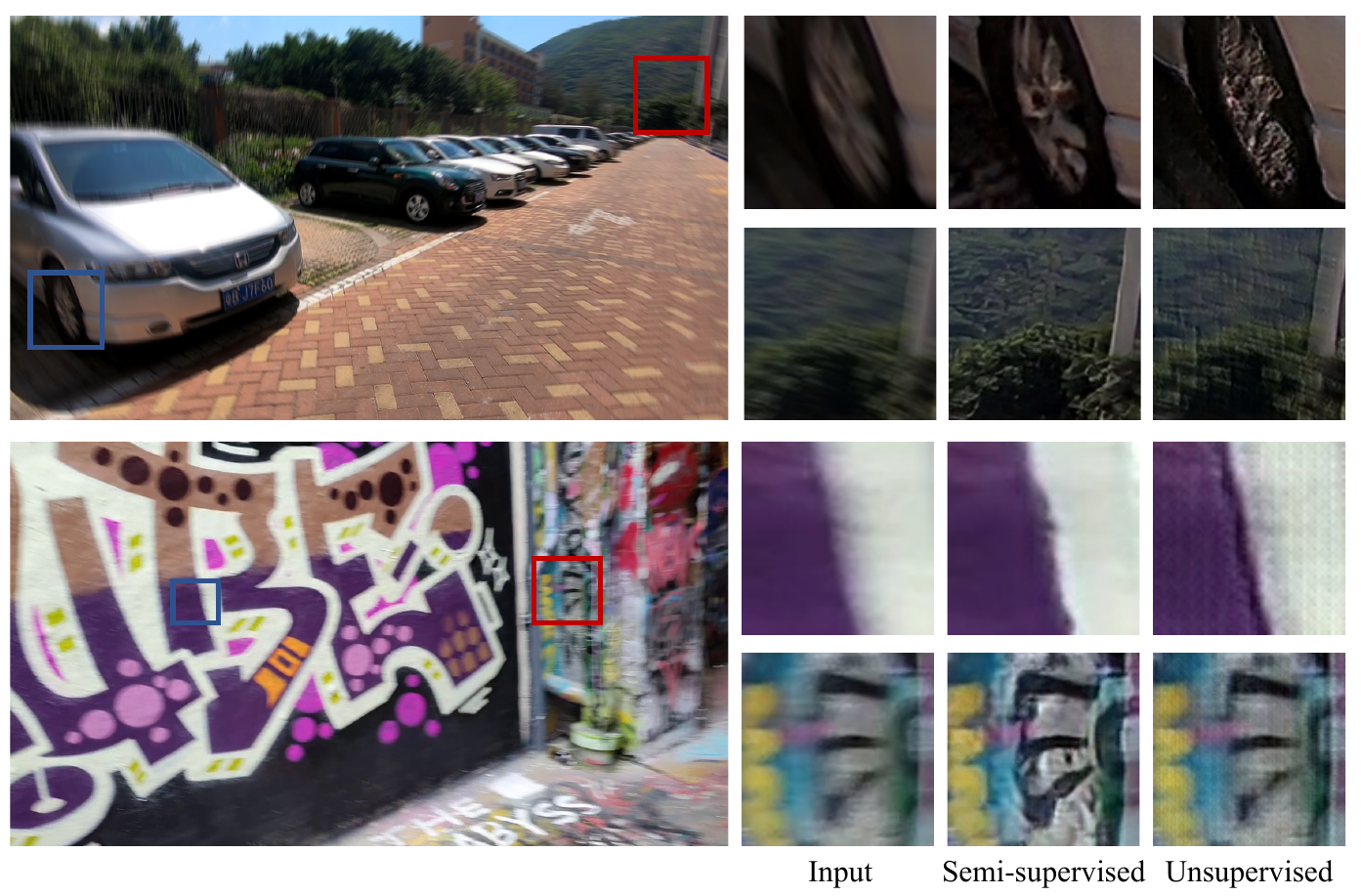}
    \caption{\textbf{Results of unsupervised deblurring with NeurMAP.} The examples are from the kernel-synthesized blur test set (upper row) and the Real-world blur test set (lower row) respectively.}
    \label{fig:unsup}
\end{figure}

\noindent A: Theoretically, the proposed NeurMAP can be trained only using unpaired data. Actually, we tried this fully unsupervised training scheme, and
experiments (Fig.~\ref{fig:unsup}) show that our model is still able to preserve the content consistency in color and most texture. However, the deblurring effect is not satisfactory and different artifacts arise in kernel-synthesized data and Real-world blurry data. These suggest that our NeurMAP is still troubled by ill-posedness under unsupervised training setting. Supervised training on paired data actually performs as a regularization to our NeurMAP. Meanwhile, it ensures the quantitative score on the GoPro paired dataset. We will conduct further research for unsupervised deblurring in the future.


\section{Conclusion}
In this paper, we propose a NeurMAP framework that consists of a deblurring network, a motion estimation network and an natural image discriminator. Based on the proposed NeurMAP estimation, all these networks are trained jointly to remove the blur patterns within unpaired data. Experiments show that our NeurMAP significantly improves the generalization ability to unseen blurriness. However, there still exist challenges, \textit{e.g.}  still requiring paired data during training. In future works, We hope to explore better image priors or introduce high-level semantic guidance to solve these problems.


%

\ifCLASSOPTIONcompsoc

\ifCLASSOPTIONcaptionsoff
  \newpage
\fi



\bibliographystyle{IEEEtran}
\bibliography{egbib_abbv}
\end{document}